\documentclass[sigconf]{acmart}

\AtBeginDocument{%
  }

\setcopyright{acmlicensed}
\copyrightyear{2025}
\acmYear{2025}
\setcopyright{cc}
\setcctype{by}
\acmConference[ICMI '25]{Proceedings of the 27th International Conference on Multimodal Interaction}{October 13--17, 2025}{Canberra, ACT, Australia}
\acmBooktitle{Proceedings of the 27th International Conference on Multimodal Interaction (ICMI '25), October 13--17, 2025, Canberra, ACT, Australia}
\acmDOI{10.1145/3716553.3750775}
\acmISBN{979-8-4007-1499-3/2025/10}

\usepackage{xcolor}
\usepackage{multirow}
\usepackage{xspace}          
\newcommand{\modelname}{WatchHAR\xspace}

\begin{document}

\title{\modelname{}: Real-time On-device Human Activity Recognition System for Smartwatches}

\author{Taeyoung Yeon}
\orcid{0009-0004-5552-0764}
\affiliation{%
  \institution{Northwestern University}
  \city{Evanston}
  \state{IL}
  \country{USA}
}
\email{taeyoungyeon@northwestern.edu}

\author{Vasco Xu}
\orcid{0000-0003-3990-582X}
\affiliation{%
  \institution{University of Chicago}
  \city{Chicago}
  \state{IL}
  \country{USA}
}
\email{vascoxu@uchicago.edu}

\author{Henry Hoffmann}
\orcid{0000-0003-0816-8150}
\affiliation{%
  \institution{University of Chicago}
  \city{Chicago}
  \state{IL}
  \country{USA}
}
\email{hankhoffmann@cs.uchicago.edu}  

\author{Karan Ahuja}
\orcid{0000-0003-2497-0530}
\affiliation{%
  \institution{Northwestern University}
  \city{Evanston}
  \state{IL}
  \country{USA}
}
\email{kahuja@northwestern.edu}

\renewcommand{\shortauthors}{Yeon et al.}
\renewcommand{\shortauthors}{Yeon et al.}
\newcommand{\todo}[1]{\textcolor{red}{[TODO: #1]}}
\newcommand{\name}{{\modelname{}}}

\begin{abstract}
Despite advances in practical and multimodal fine-grained Human Activity Recognition (HAR), a system that runs entirely on smartwatches in unconstrained environments remains elusive. We present \modelname{}, an audio and inertial-based HAR system that operates fully on smartwatches, addressing privacy and latency issues associated with external data processing. By optimizing each component of the pipeline, \modelname{} achieves compounding performance gains. We introduce a novel architecture that unifies sensor data preprocessing and inference into an end-to-end trainable module, achieving 5x faster processing while maintaining over 90\% accuracy across more than 25 activity classes. \modelname{} outperforms state-of-the-art models for event detection and activity classification while running directly on the smartwatch, achieving 9.3\,ms processing time for activity event detection and 11.8\,ms for multimodal activity classification. This research advances on-device activity recognition, realizing smartwatches' potential as standalone, privacy-aware, and minimally-invasive continuous activity tracking devices.
\end{abstract}


\begin{teaserfigure}
  \includegraphics[width=\columnwidth]{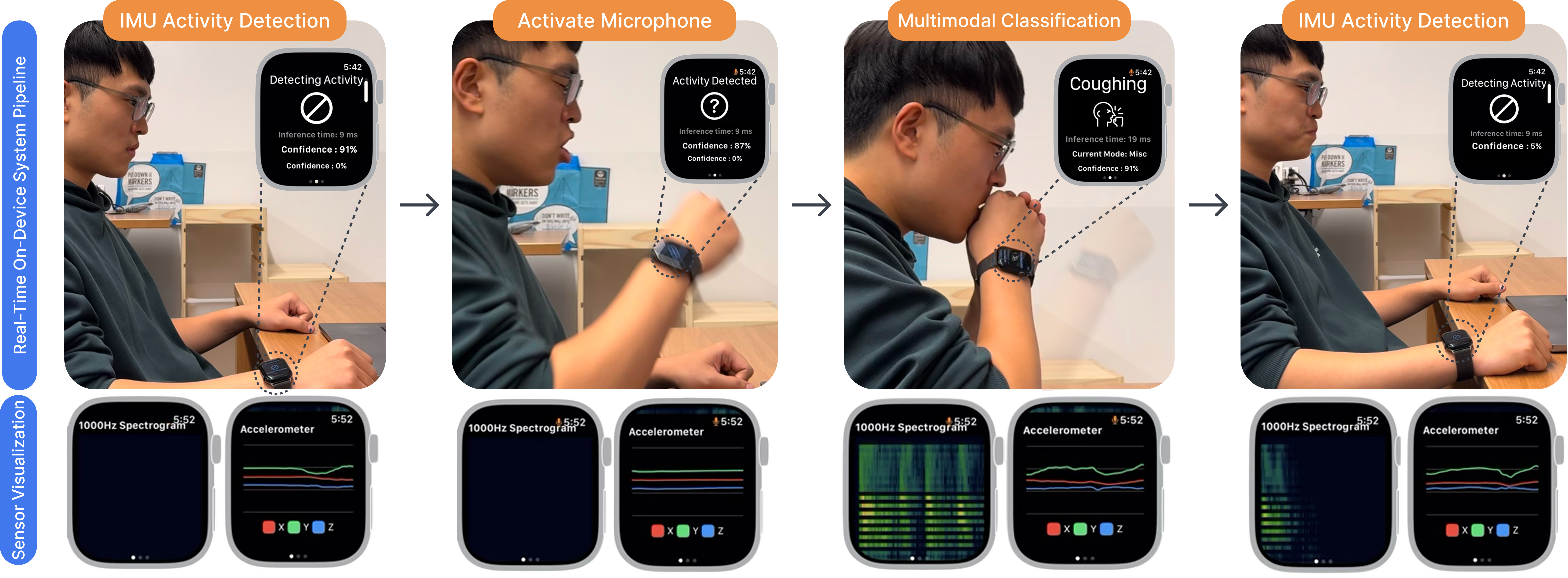}
  \caption{Our application runs in real-time directly on a smartwatch, detecting and classifying human activities. The system first uses IMU data to detect the presence of an event. Once an event is detected, it triggers a multimodal activity classifier that activates the microphones and uses both IMU and audio data to classify the activity. }
  \Description{Our application runs in real-time directly on a smartwatch, detecting and classifying human activities. The system first uses IMU data to detect the presence of an event. Once an event is detected, it triggers a multimodal activity classifier that activates the microphones and uses both IMU and audio data to classify the activity. }
  \label{fig:teaser}
\end{teaserfigure}


\maketitle

\section{Introduction}

Human Activity Recognition (HAR) has become a cornerstone of ubiquitous computing, with applications ranging from health monitoring and context-aware services to assistive technologies for people who are deaf or hard of hearing. While significant strides have been made in developing accurate and robust HAR systems for Activities of Daily Living (ADL), a persistent challenge has been creating solutions that are both practical for everyday use and capable of operating in unconstrained environments. Smartwatches, with their array of sensors and constant proximity to users, present an ideal platform for HAR. However, most current smartwatch-based sensing systems rely on external data processing, raising concerns about privacy, latency, and the need for constant connectivity to remote machines to offload compute \cite{samosa, bhattacharya2022leveraging, prismtracker, ahuja2024practical}. 

On-device processing directly addresses these concerns by ensuring real-time feedback and safeguarding sensitive information. For example, workers in construction or manufacturing often operate in areas with limited connectivity; delayed notifications due to reliance on remote servers can undermine safety or productivity. Similarly, elder-care systems require continuous tracking to detect emergencies in real-time. In such cases, offloading sensitive health data to external infrastructure not only risks privacy breaches but also increases response latency.

\begin{figure}[t]
  \centering
  \includegraphics[width=\columnwidth]{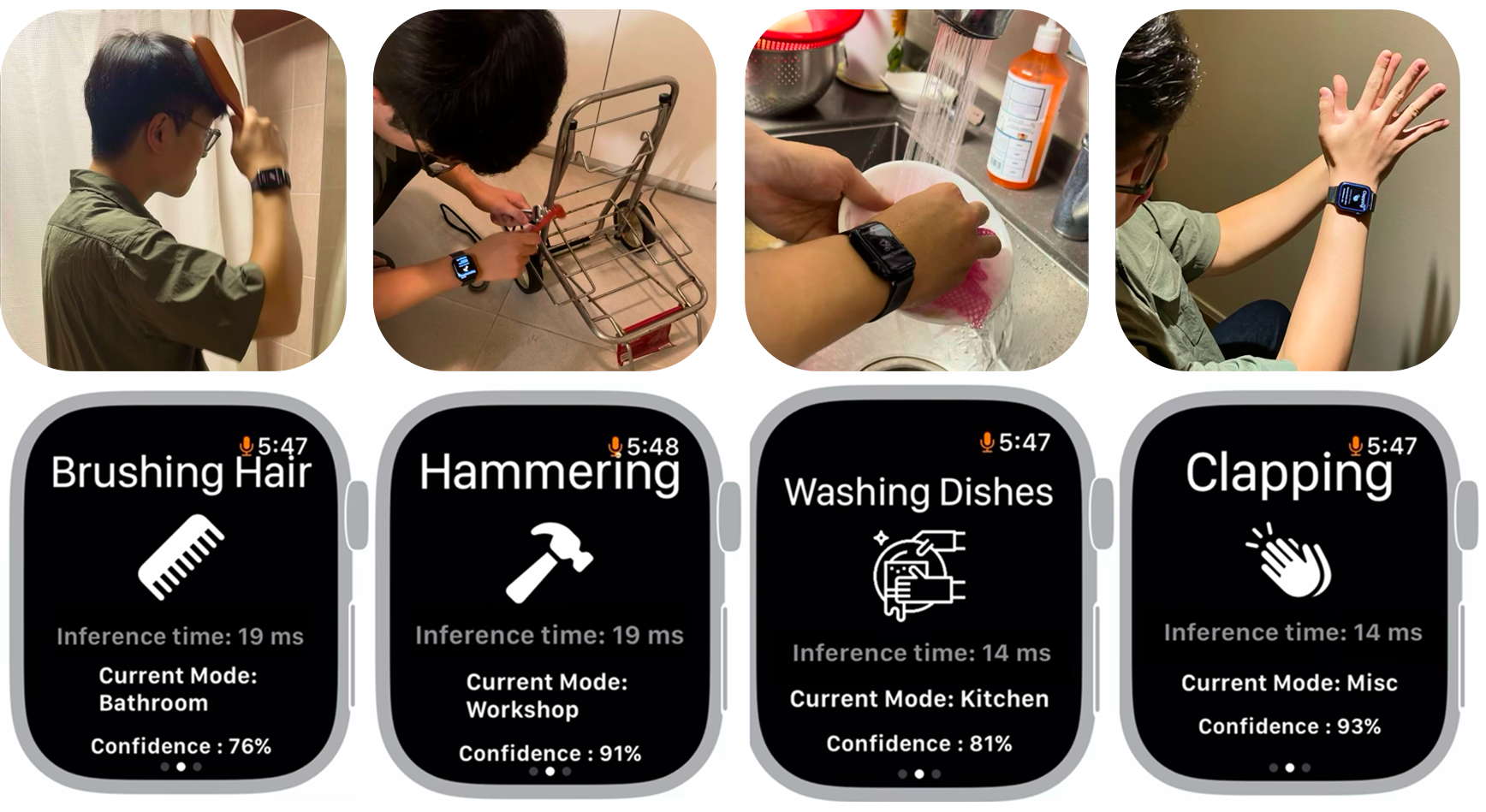}
  \caption{Our \modelname{} system running in real-time on an Apple Watch Series 7 (45mm), demonstrating activity recognition across four different contexts: brushing hair, hammering, washing dishes, and clapping.}
  \Description{Our Application is running at real-time smartwatch on-device with different human activities.}
  \label{fig:contextwise}
\end{figure}

\modelname{} addresses these challenges by introducing a novel HAR system that operates entirely on smartwatches, leveraging both audio and inertial data (Figure \ref{fig:contextwise}). We chose IMU and audio sensors as they are universally available on smartwatches and provide complementary information -- audio provides distinctive signatures for audible activities but is power-intensive, while IMUs are lightweight and can capture fine-grained hand movements but produce less distinct signals. By eliminating the need for external data processing, \modelname{} enhances privacy and reduces latency while maintaining high accuracy across a wide range of activities. The system employs a two-stage approach: a lightweight IMU-based activity detector that triggers a more resource-intensive multimodal classifier only when necessary. This strategy optimizes power consumption without sacrificing performance (Section \ref{sec:imu_event_detector}).
This is enabled by \modelname{}'s end-to-end trainable preprocessing module, which applies a Short-Time Fourier Transform (STFT) and approximates a mel-filter bank as a 1D convolutional operation that runs efficiently on mobile neural processors.

Through careful optimization of each component of the pipeline, \modelname{} achieves compounding performance gains. The system outperforms the latest models for event detection (by 5.5\%, Table \ref{tab:event_detection_comparison}) and even achieves modest improvements in activity classification (by 0.7\%) while running entirely on the smartwatch, with processing times of 9.3\,ms for activity detection and 11.8\,ms for multimodal classification. \modelname{}’s novel architecture unifies sensor data preprocessing and inference into a single, trainable module, providing a 5x performance boost while maintaining over 90\% accuracy across more than 25 activity classes. These advancements demonstrate \modelname{}’s potential to revolutionize on-device activity recognition, realizing the full potential of smartwatches as standalone devices with minimally invasive activity tracking.

Our main contributions are:

(1) A complete multimodal HAR system running entirely on commodity smartwatches with real-time performance (9.3\,ms event detection, 11.8\,ms activity classification);

(2) An end-to-end trainable audio preprocessing module integrating STFT and mel-filter banks directly into the neural network for efficient on-device execution;

(3) Comprehensive evaluation demonstrating 5× faster processing and 5-47× lower computational cost (FLOPs) than state-of-the-art while maintaining competitive accuracy;

(4) Open-source implementation, models, and evaluation scripts to foster reproducibility and community adoption \footnote{\url{https://github.com/SPICExLAB/WatchHAR}}.

\section{Related Work}

Human Activity Recognition (HAR) has seen significant advancements in recent years, particularly in the domain of wearable technology. A wide range of wearable devices have been explored for HAR, including wrist-worn sensors \cite{morris2014recofit, akther2021mteeth, wristwash, activityposer}, smart rings \cite{washring}, earbuds \cite{2022coughtrigger}, and smartwatches \cite{samosa, kunwar2022robust, bhattacharya2022leveraging, 2019swimmingstyle, 2016Tapskin}. These devices have proven effective in identifying various activities, from fitness exercises to daily tasks. 

Smartwatches, housing a rich collection of sensors including IMUs and microphones, have proven highly effective for activity recognition in everyday settings. For example, SAMoSA \cite{samosa} and Bhattacharya et al. \cite{bhattacharya2022leveraging} highlight the benefits of combining audio and IMU data. SAMoSA achieved 92.2\% accuracy across various contexts using 1kHz audio and 50Hz IMU data, demonstrating that even lower-sampled audio can enhance activity recognition while preserving privacy. They proposed an IMU-based activity detector that activates the microphone only upon detecting an activity, enabling efficient multimodal classification. Bhattacharya et al. \cite{bhattacharya2022leveraging} showcased robust multimodal sensor fusion techniques, performing well in both controlled and real-world scenarios. Despite these advancements, most systems treat smartwatches merely as data collectors and offload processing to smartphones or desktops \cite{2016Tapskin, washring, samosa, bhattacharya2022leveraging}. This approach, while computationally effective, compromises privacy and real-time responsiveness. 

Some efforts have been made towards on-device processing, such as Kim et al.'s \cite{kim2023proxifit} exercise monitoring system using natural magnetism in exercise equipment, and Zhang et al.'s \cite{2022coughtrigger} cough detection system that uses IMU sensor values to activate cough detection. Kunwar et al. \cite{kunwar2022robust} also explored robust and deployable gesture recognition for smartwatches. However, these solutions primarily target a limited range of classes and utilize IMU data, avoiding the power-hungry and computationally intensive audio processing despite its rich contextual information. 

The key challenge lies in developing a system that can leverage the rich information from both audio and IMU sensors to support the fidelity of HAR while operating entirely on resource-constrained wearable devices. This requires not only efficient algorithms but also novel approaches to sensor data processing, gating and fusion. \modelname{} overcomes these challenges by implementing a 1D convolution approach for generating log-mel spectrograms and combining it with efficient convolutional classifier architectures, allowing the model to run on smartwatch neural accelerators in real-time. 

\begin{figure*}
  \centering  
\includegraphics[width=\textwidth,keepaspectratio]{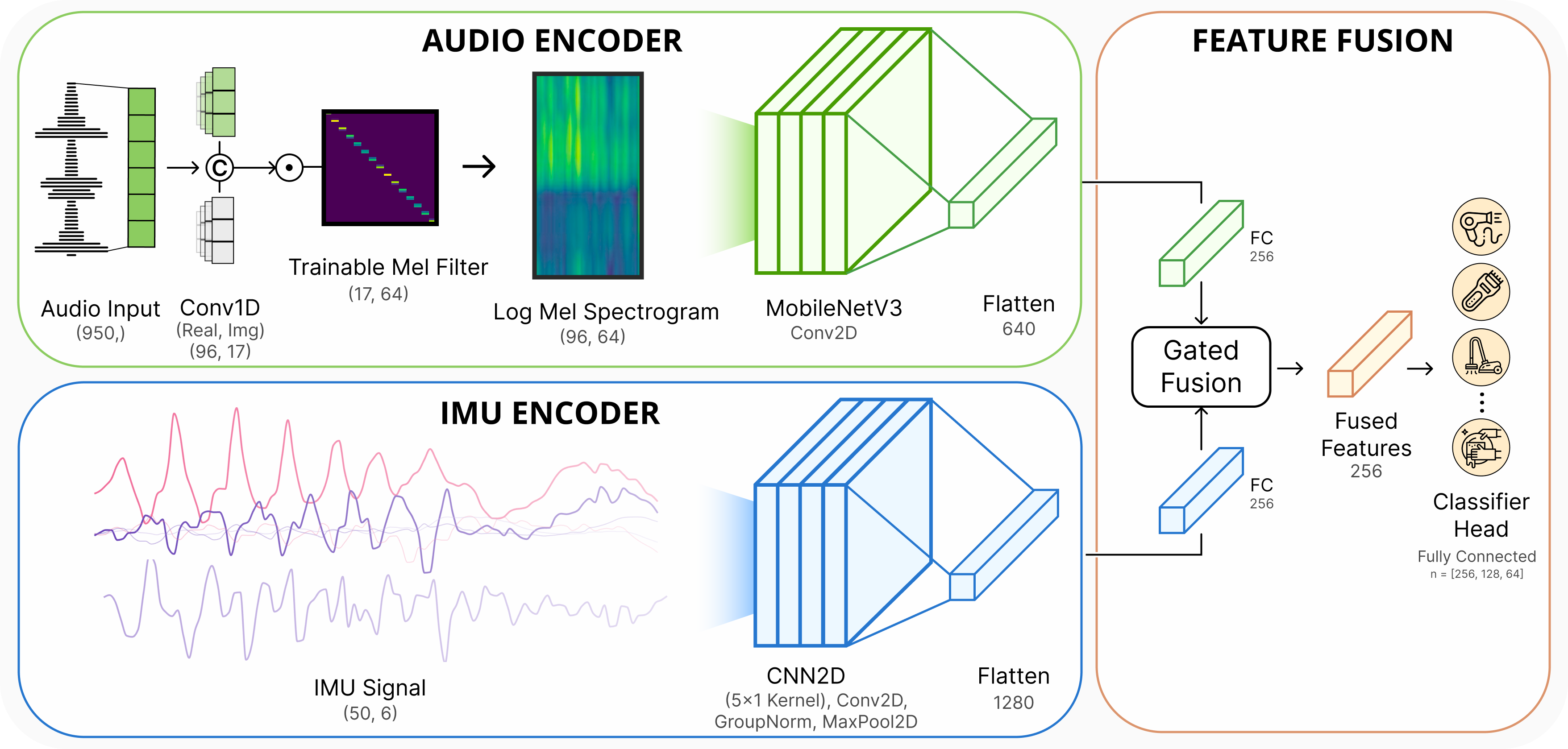}
  \caption{Overview of \modelname{}'s Multimodal Activity Classifier. The model processes 1-second windows of raw audio and 6-axis IMU data to predict activities. Audio preprocessing is integrated directly into the neural network as trainable layers.}
  \Description{Overall architecture of \modelname{}'s Multimodal Activity Classifier}
  \label{fig:architecture}
\end{figure*}

\section{Model Architecture}

Building upon prior work, SAMoSA\cite{samosa}, which introduced the concepts of the IMU Event Detector and Multimodal Activity Classifier, we have significantly improved the model architecture to optimize for on-device application. Our system architecture maintains these two core components while balancing computational efficiency and power consumption for real-time activity recognition on smartwatches (Figure \ref{fig:architecture}).

To ensure our system runs efficiently on smartwatches, we implement several optimization techniques on the trained models. We first traced our PyTorch models using torch.jit.trace with example tensors matching our expected input shapes with batch size 1 for simulating real-time inference. This intermediate representation was then converted to CoreML \cite{CoreMLApple} for optimized execution on Apple Watch hardware. We apply 16-bit float quantization, which halves the model size and improves inference speed with no impact on accuracy. All inference runs on the Apple Watch  Series 7 (45mm) GPU via CoreML for optimal performance. We carefully tune the window sizes and hop lengths for both the event detector and classifier to balance between accuracy, latency, and computational load. These optimizations enable our system to run in real-time on commodity smartwatch hardware while maintaining high accuracy across a wide range of activities.

\subsection{IMU-Only Event Detector}
\label{sec:imu_event_detector}

We use a lightweight IMU-based event detector to trigger the more resource-intensive multimodal classifier. Our detector uses a 1D depthwise Convolutional Neural Network (CNN) architecture \cite{chollet2017xception}, processing 3-second windows of 6-axis IMU data (3-axis accelerometer and 3-axis gyroscope) sampled at 50 Hz. The model consists of four convolutional blocks with increasing filter counts (64 to 128) and decreasing kernel sizes (10 to 5), interspersed with max pooling layers, followed by fully connected layers (512, 256, 128 nodes) and a final sigmoid output for binary event detection (activity detected or not). 

To ensure rapid detection of event onset while minimizing false positives from small motions, we use a 3-second rolling window with a 20 ms hop length and apply a 2-second moving average to filter out spurious detections. Our system uses a two-stage detection process to balance performance and power consumption. The IMU-Only Event Detector continuously monitors for the presence of an event and triggers the Multimodal Activity Classifier, activating the microphone, only when an event is detected. Otherwise, the microphone remains off, conserving energy during idle periods.

\subsection{Multimodal Activity Classification}
\label{sec:multimodal_classification}

Our Multimodal Activity Classifier (Figure~\ref{fig:architecture}) processes both IMU and audio data to achieve high-accuracy activity recognition. Since audio adds a significant computational overhead, we use shorter window sizes to enable faster processing and reduce latency. Both IMU and audio data use 1-second windows with a 20 ms hop length, allowing for fine-grained temporal resolution in our classifications.

We implement an end-to-end trainable audio preprocessing module directly within our neural network \cite{nnaudio}. Our architecture consists of three main components: a Short-Time Fourier Transform (STFT) implemented using a 1D convolutional layer, a mel-filter bank designed as a trainable linear layer, and an amplitude-to-DB conversion using a logarithmic activation function. The STFT layer uses two separate convolutions for the real and imaginary parts, with kernel size corresponding to the FFT size and stride determining the hop length. The mel-filter bank layer is initialized with triangular mel filters but remains trainable, potentially learning optimized filter banks for our specific human activity recognition task. This is key, as filters suited for HAR may differ greatly from those originally designed for speech recognition tasks. Lastly, the amplitude-to-DB conversion uses a logarithmic activation function to produce a spectrogram that is then passed to the audio feature encoder.

For the audio encoder, we use a MobileNetV3 \cite{howard2019mobilenetv3} backbone pretrained on the AudioSet dataset \cite{audioset}, without any platform-specific modifications. This choice offers a good balance between model size, computational efficiency, and accuracy. The MobileNetV3 architecture incorporates inverted residual blocks with squeeze-and-excitation modules, platform-aware neural architecture search for optimized layer design, and efficient last-stage design for classification tasks.

For the IMU encoder, we adopt the ConvBoost architecture \cite{convboost}. This model uses a standard 3-layer 2D CNN structure optimized for efficient processing of multivariate time series data such as IMU signals. Each layer uses 5×1 kernels to extract temporal features, with max pooling applied after the first two convolutional layers. ReLU activations are used throughout the network for non-linearity. The classifier consists of two fully connected layers with dropout ($p = 0.5$) regularization to prevent overfitting. This architecture prioritizes simplicity and computational efficiency while maintaining effective feature extraction for human activity recognition tasks.

To effectively combine information from IMU and audio modalities, we implement a Gated Fusion mechanism \cite{arevalo2017gatedmultimodalunitsinformation} rather than simple feature concatenation. Both IMU and audio embeddings are first projected into a shared 256-dimensional space using separate linear layers. Each projected embedding is then passed through its own gating network -- a linear layer followed by a sigmoid activation  -- that produces a vector of weights between 0 and 1. These weights are applied to the projected features through element-wise multiplication, effectively scaling each feature dimension by its learned importance. The gated IMU and audio features are then summed together and passed through an additional linear layer to produce the final 256-dimensional fused representation. This fused vector is then passed through a final classifier head to predict class probabilities. All components are fully differentiable and jointly trained with the rest of the model.

The gated fusion mechanism is trained end-to-end with binary cross-entropy loss, learning to weight modalities based on their relevance to each activity. This approach is particularly effective in the P-LOPO setting (Section \ref{sec:ablation}) where some personalization data is available for user adaptation. 
\begin{table*}[!t]
\centering
\caption{Event Detection Model Performance Comparison between SAMoSA and \modelname{} on Apple Watch Series 7}

\resizebox{\textwidth}{!}{%
\begin{tabular}{lccccc}
    \toprule
    \textbf{Method} & \textbf{F1 score (\%)} & \textbf{Processing Time (ms)} & \textbf{Onset Latency (sec)} & \textbf{Offset Latency (sec)} \\ 
    \midrule
    \textbf{SAMoSA [IMU @ 50 Hz]} & 88.0 & 55.4 & 0.62 & 0.16 \\ 
    \textbf{\modelname{} [IMU @ 50 Hz]} & \textbf{93.5} & \textbf{9.3} & \textbf{0.27} & \textbf{0.07} \\ 
    \bottomrule
\end{tabular}
} 

\label{tab:event_detection_comparison}
\end{table*}

\section{Datasets}

To train and evaluate \modelname{}, we utilize the following publicly available smartwatch datasets, which were collected in previous studies.

\textbf{SAMoSA Dataset \cite{samosa}:} The SAMoSA dataset was collected from 20 participants (mean age 23.3, all right-handed) across 60 diverse environments. Data was recorded using a Fossil Gen 5 smartwatch, capturing synchronized 9-axis IMU data (accelerometer, gyroscope, and orientation) at 50 Hz and uncompressed audio at 16 kHz, later post-processed to 1 kHz for privacy preservation. The dataset includes 26 activities performed in four contexts: kitchen, bathroom, workshop, and miscellaneous. Each participant performed every activity three times per context, resulting in 14.2 hours of data in total -- 5.9 hours of labeled activity data and 8.3 hours of transition (``Other") data. All activities were performed in participants' homes using their own appliances and tools, naturally incorporating ambient background noise. This dataset is used to train and evaluate both activity detection and classification.

\textbf{Semi-Naturalistic and In-the-Wild Dataset \cite{bhattacharya2022leveraging}:} 
We used two complementary datasets from Bhattacharya et al.’s work. First, the \textbf{Semi-Naturalistic dataset} was collected from 15 participants (9 female, 6 male, mean age 43.6), representing diverse professional and socioeconomic backgrounds. Data was captured using a Fossil Gen 4 smartwatch, recording accelerometer and gyroscope data at 50 Hz, and audio data at 22.05 kHz. Participants performed 23 activities twice across two sessions, with each activity lasting a minimum of 30 seconds. Data collection was conducted remotely via video calls, and participants knocked on a surface to mark the start and end of activities. Continuous recordings captured all activities, including in-between movements, and manual annotation was performed using sensor data and video footage.

In addition, an \textbf{In-the-Wild dataset} was gathered from five additional participants (4 males, 1 female, mean age 27). These participants wore the same smartwatch alongside a smartphone mounted on their chest, which captured 25-second egocentric video clips every minute using a dedicated mobile application. Data collection for the in-the-wild study was performed without any predefined activity scripts, allowing participants to engage in their daily routines naturally over two separate sessions each, totaling 10 in-the-wild sessions. 
\section{Evaluation}

We evaluate \modelname{}'s performance against prior smartwatch-based approaches, focusing on aspects critical for real-time applications: processing time, model size, and accuracy across different settings. All models were implemented using PyTorch version 2.1.2 and converted to CoreML format using coremltools version 7.1 with float16 quantization for running on Apple Devices. Converting 32-bit models to 16-bit had no impact on accuracy while halving the model size. On-device performance evaluations were conducted using an Apple Watch Series 7 (45mm) GPU via CoreML library.

\subsection{IMU-Only Event Detector}

We compare our Depthwise CNN1D Event Detection Model with SAMoSA's Random Forest event detection model, as shown in Table \ref{tab:event_detection_comparison}. \modelname{} outperforms SAMoSA across all metrics, achieving higher F1 scores (93.5\% vs. 88.0\%). As noted in Section~\ref{sec:imu_event_detector}, to address potential mispredictions due to data skewness, we applied a 2-second moving average to the output probabilities, smoothing predictions and reducing spurious outputs. This improved our F1 score from 92.5\% to 93.5\% with negligible computational overhead.  

Compared to SAMoSA, \modelname{} demonstrates significantly faster processing times (9.3 ms vs. 55.4 ms)\footnote{SAMoSA originally reported 4.8 ms on MacBook Air (M1). We measured both methods on Apple Watch Series 7 (45mm) for fair comparison. WatchHAR's event detector inference time is 2.1 ms on M1 hardware}. Processing time includes data preprocessing and model inference for a single IMU window. \modelname{}'s processing efficiency stems from streamlined GPU feature computation. SAMoSA calculates eight statistical features -- mean, standard deviation, max, min, median, variance, skewness, and kurtosis -- for each of the nine IMU values, requiring 28.76 ms on the Apple Watch Series 7 (45mm). In contrast, our 1D CNN only normalizes raw IMU data, completing in just 3 ms. We also measured onset latency (i.e., the delay between the physical start of an event and its detection by the model) and offset latency (i.e., the delay in detecting the end of an event). \modelname{} achieves lower onset (0.27s vs 0.62s) and offset (0.07s vs 0.16s) latencies compared to SAMoSA.

\begin{table*}[ht]
\centering
\caption{Multimodal Activity Classification across different approaches and datasets.
For SAMoSA \& Semi-Naturalistic, their evaluation metrics are LOPO or P-LOPO. Their values are classification accuracies (\%). Note, for In-the-Wild dataset, their evaluation metrics are weighted F1 scores.}
\resizebox{\textwidth}{!}{%
\begin{tabular}{llccccc}
\toprule
\textbf{Model Name} & 
\textbf{Dataset} & 
\textbf{Sampling Rate (kHz)} & 
\textbf{LOPO} & 
\textbf{P-LOPO} & 
\textbf{Processing Time (ms)} & 
\textbf{FLOPs (G)} \\
\midrule
\textbf{SAMoSA} 
  & SAMoSA 
  & 1 
  & 92.2 
  & N/A 
  & 56.4 
  & 1.71 \\
\textbf{\modelname{}} 
  & SAMoSA 
  & 1 
  & \textbf{92.34} 
  & N/A 
  & \textbf{11.8} 
  & \textbf{0.036} \\
\midrule
\textbf{Bhattacharya et al.} 
  & Semi-Naturalistic 
  & 22.05 
  & 89.7 
  & \textbf{94.3} 
  & 438.3 
  & 4.24 \\
\textbf{\modelname{}} 
  & Semi-Naturalistic 
  & 22.05 
  & \textbf{90.4} 
  & 93.8 
  & \textbf{71.1}
  & \textbf{0.917} \\
\midrule
\textbf{Bhattacharya et al.} 
  & In-the-Wild 
  & 22.05 
  & N/A 
  & 55.8
  & 438.3 
  & 4.24 \\
\textbf{\modelname{}} 
  & In-the-Wild 
  & 22.05 
  & N/A
  & \textbf{56.7} 
  & \textbf{71.1} 
  & \textbf{0.917} \\
\bottomrule
\end{tabular}
}
\label{tab:activity_comparison}
\end{table*}

\subsection{Multimodal Activity Classification}
\label{sec:multimodal_results}

We compare our Multimodal Activity Classifier with two prior works, SAMoSA \cite{samosa} and Bhattacharya et al. \cite{bhattacharya2022leveraging} on their respective datasets, as summarized in Table \ref{tab:activity_comparison}. All models were evaluated using a leave-one-participant-out (LOPO) cross-validation scheme or Personalized-LOPO (P-LOPO), which incorporates a subset of test participants' data during training to simulate partial personalization. In our P-LOPO evaluation, we use personalized data from public datasets, though future work could explore on-device fine-tuning to preserve user privacy. For the In-the-Wild dataset, we report weighted-F1 scores evaluated in a P-LOPO protocol, aligning with the evaluation protocol from Bhattacharya et al. \cite{bhattacharya2022leveraging}.  We also compare processing times, measured on an Apple Watch Series 7 (45mm), which reflects the total time to process a single window of audio and IMU data, from log-mel generation to prediction.

On the SAMoSA dataset, we compare with the authors' pretrained model configured for 1 kHz audio and 50 Hz motion data, matching their primary evaluation setup. We compute the context-wise accuracy defined as the average classification accuracy across four contexts: kitchen, bathroom, workshop, and miscellaneous. \modelname{} achieves slightly higher (92.34\% vs 92.2\%) context-wise accuracy with $47\times$ lower computational cost (0.036 GFLOPs vs 1.71 GFLOPs) and $5\times$ faster processing time (11.8 ms vs 56.4 ms).

On the Semi-Naturalistic dataset, we reproduced the best performing architecture from their paper. CNN14 \cite{kong2020pannslargescalepretrainedaudio} for audio and \linebreak Attend\&Discriminate \cite{attend_discriminate} for IMU with concatenation late fusion method, since no pretrained model was provided. \modelname{} achieves slightly better LOPO accuracy (90.4\% vs 89.7\%) and comparable P-LOPO accuracy (93.8\% vs 94.3\%), while requiring $11\times$ fewer FLOPs (0.917 GFLOPs vs 4.24 GFLOPs) and achieving $6\times$ faster inference time (71.1 ms vs 438.3 ms). Note that our FLOPs increased from the SAMoSA dataset (0.036 GFLOPs) to the Semi-Naturalistic dataset (0.917 GFLOPs) due to adjusting our CNN2D IMU encoder to handle the longer 10-second input windows compared to 1-second windows used for SAMoSA dataset.

On the In-the-Wild dataset, we followed Bhattacharya et al.’s evaluation protocol: \modelname{} was first pretrained using the Semi-Naturalistic dataset and then fine-tuned with within-session data, leading to a personalized leave-one-participant-out (P-LOPO) evaluation. Similar to  Bhattacharya et al., we augmented the training data with Semi-Naturalistic samples to address class imbalance and missing labels. \modelname{} achieves higher weighted F1 scores compared to Bhattacharya et al. (56.7\% vs 55.8\%). This drop in accuracy -- relative to those achieved on the Semi-Naturalistic dataset (over 90\%) -- are expected, as the In-the-Wild dataset contains noisier ground truth labels due to limitations in the video-based annotation system used during data collection. Despite these challenges, our system achieves a higher F1 score while requiring $11\times$ lower computational cost. \modelname{}, without any per-user fine-tuning and using only the model trained on the Semi-Naturalistic dataset, obtains a weighted F1 score of 28.5 compared to 26.8 from Bhattacharya et al. \cite{bhattacharya2022leveraging}.

We refer readers to the Appendix for per-activity confusion matrices on the SAMoSA, Semi-Naturalistic, and In-the-Wild dataset across each activity context. All our models, code, and evaluation scripts are publicly available at \url{https://github.com/SPICExLAB/WatchHAR} to foster community use and adoption.

\section{Ablation Study}
\label{sec:ablation}

We analyze how different architectural choices affect the accuracy and computational complexity through an ablation on the \textit{Semi‑Naturalistic} dataset. We explore performance-efficiency tradeoffs across different pretrained audio backbone models, IMU encoder architectures, and multimodal feature fusion techniques. Model variants are trained and evaluated using the protocols outlined in Sections \ref{sec:multimodal_classification} and \ref{sec:multimodal_results}, respectively. Additionally, we report model size and computational cost (FLOPs) to assess their suitability for on-device applications.


We evaluated four publicly available pretrained audio encoders — CNN14, ResNet-22, MobileNetV1, and our choice of MobileNetV3 — all pretrained on AudioSet~\cite{kong2020pannslargescalepretrainedaudio}. Of note, we excluded the pretrained VGGish model used in SAMoSA as it is hard-coded for $96{\times}64$ spectrogram inputs, making it incompatible with the $690{\times}64$ dimensions of the \textit{Semi-Naturalistic} dataset. As Table \ref{table:audio_ablation} shows, ResNet‑22 yields the highest LOPO score (79.8\%), but MobileNetV3 attains the best P‑LOPO (86.7\%) while being $\scriptstyle\sim$30$\times$ smaller and $\scriptstyle\sim$13$\times$ lighter in FLOPs.


We also evaluated four IMU backbone architectures: CNN1D from SAMoSA \cite{samosa}, DeepConvLSTM \cite{convboost}, Attend\&Discriminate \cite{attend_discriminate}, and our choice of CNN2D \cite{convboost}. Unlike the audio encoders, all IMU encoders were initialized from scratch. As shown in Table \ref{tab:imu_ablation}, Attend\&Discriminate achieved the highest P-LOPO accuracy (90.8\%), consistent with prior results from Bhattacharya et al. \cite{bhattacharya2022leveraging}, while our CNN2D achieved the best LOPO performance (85.5\%) with the lowest computational cost (0.13 GFLOPs).


Finally, we reproduced the three late‑fusion schemes proposed by Bhattacharya et al.\cite{bhattacharya2022leveraging}: (1) simple concatenation of modality embeddings, (2) cross-modal self-attention, and (3) softmax averaging of independent classifications. We compare these schemes with our gated‑fusion method, which adds small gated block that learns per‑sample modality weights. As shown in Table \ref{tab:fusion_ablation}, softmax averaging achieved the highest LOPO accuracy (90.5\%) but lower P-LOPO performance (92.8\%). Our gated-fusion approach nearly matched the LOPO score while improving P-LOPO accuracy to 93.8\%, with minimal computational overhead (<0.5M parameters, negligible FLOPs increase). These results demonstrate that our fusion method particularly excels when user-specific data is available for fine-tuning.



\begin{table}[h]
\centering
\begin{tabular}{lcccc}
\hline
\textbf{Model} & \textbf{LOPO} & \textbf{P‑LOPO} & \textbf{Params (M)} & \textbf{FLOPs (G)} \\ \hline
CNN14           & 74.5 & 82.4 & 80.80 & 14.54 \\ \hline
ResNet-22        & \textbf{79.8} & 86.5 & 63.73 & 10.36 \\ \hline
MobileNetV1     & 77.8 & 85.5 & 5.36  & 1.25  \\ \hline
MobileNetV3            & 77.8 &\textbf{86.7} & \textbf{2.19}  & \textbf{0.79}  \\ \hline
\end{tabular}
\caption{Audio Model Ablation Study Results.}
\label{table:audio_ablation}
\end{table}

\begin{table}[h]
\centering
\resizebox{\linewidth}{!}{%
\begin{tabular}{lcccc}
\hline
\textbf{Model} & \textbf{LOPO} & \textbf{P‑LOPO} & \textbf{Params (M)} & \textbf{FLOPs (G)} \\ \hline
DeepConvLSTM         & 72.0 & 77.1 & \textbf{0.72}   & 0.19 \\ \hline
CNN1D                & 75.9 & 80.4 & 246.80 & 0.56 \\ \hline
Attend\&Discriminate & 84.0 & \textbf{90.8} & 0.97   & 0.35 \\ \hline
CNN2D                & \textbf{85.5} & 89.0 & 4.57   & \textbf{0.13} \\ \hline
\end{tabular}
}
\caption{IMU Model Ablation Study Results.}
\label{tab:imu_ablation}
\end{table}

\begin{table}[h]
\centering
\resizebox{\linewidth}{!}{%
\begin{tabular}{lcccc}
\hline
\textbf{Fusion Method} & \textbf{LOPO} & \textbf{P‑LOPO} & \textbf{Params (M)} & \textbf{FLOPs (G)} \\ \hline
Concatenation      & 89.4 & 93.2 & 6.76 & 0.917 \\ \hline
Self‑Attention     & 89.8 & 93.3 & 8.53  & 0.918 \\ \hline
Softmax Averaging  & \textbf{90.5} & 92.8 & \textbf{6.76} & \textbf{0.917} \\ \hline
Gated Fusion       & 90.4 & \textbf{93.8} & 7.18  & 0.917 \\ \hline
\end{tabular}
}
\caption{Multimodal Fusion Ablation Study results.}
\label{tab:fusion_ablation}
\end{table}
\begin{figure*}
  \centering
  \includegraphics[width=\textwidth,keepaspectratio]{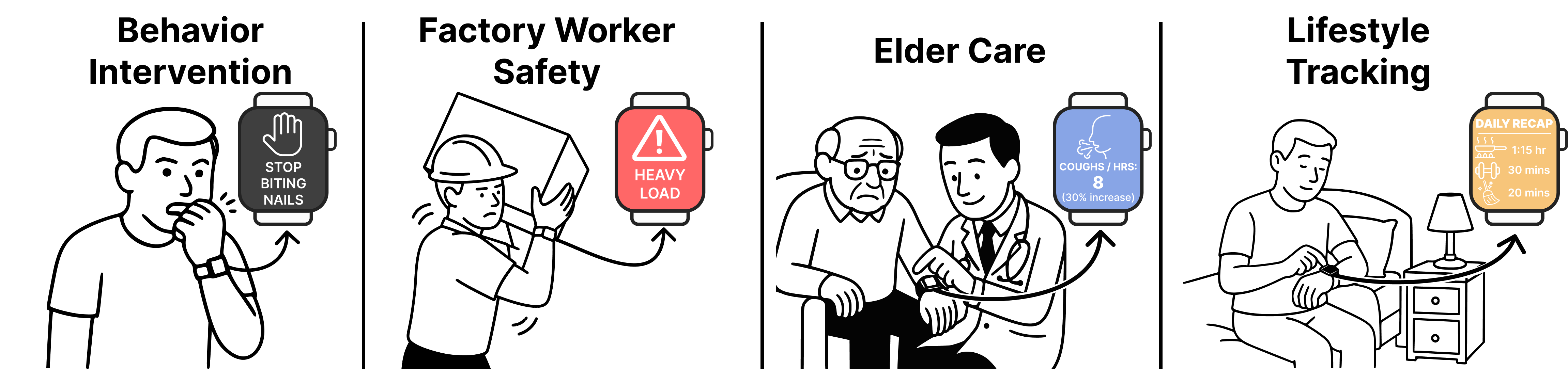}
  \caption{Illustrations of four representative application scenarios enabled by \modelname{}. 
  \textit{Left→right:} 
  Behavior Intervention—recognition of nail‑biting triggers haptic nudges; 
    Worker Safety—real‑time detection of over‑the‑shoulder lifting issues on‑the‑spot warnings; 
  Elder Care—the watch tracks cough frequency and summarizes trends for clinicians or family; 
  Lifestyle Tracking—an on‑device diary reports time spent on daily activities, giving users actionable insights without exporting raw data.}
  \Description{Four scenarios—Behavior Intervention, Factory Worker Safety, Elder Care, and Lifestyle Tracking—illustrate how on‑device activity recognition supports each domain while keeping data private.}
  \label{fig:application_scenarios}
\end{figure*}

\section{Application Scenarios}
\modelname{} broadens the scope of activity recognition by running entirely on-device, enabling real-time, privacy-preserving, and on-the-go applications across diverse application domains (Figure \ref{fig:application_scenarios}). For instance, it can be used for behavioral intervention. \modelname{} can instantly recognize gestures such as nail biting, delivering a gentle vibration, logging the event, or prompting a brief intervention - supporting real-time behavior change without external devices \cite{shoaib2015bad}.

It can also be used for worker safety in factory settings. Manual tasks in industrial settings -- such as drilling, lifting, or assembly -- produce distinctive motion and audio patterns. Early systems recognized these patterns using body-worn sensors \cite{ward2006assembly}, but required specialized hardware and offline processing. \modelname{} brings these capabilities to scale using only off-the-shelf smartwatches, enabling real-time procedure tracking, automated task logging, and alerts for high-risk movements like over-the-shoulder lifts.

\modelname{} can also be used for health sensing and elder care. By monitoring daily motion profiles \cite{xu2024mobileposer} for actions such as walking, dressing, or coughing entirely on the device, it can passively track digital health biomarkers over time. Similarly, it can be used to turn raw sensor data into a daily activity journal \cite{burke2011selfmonitor} - capturing time spent in daily activities such as cooking, cleaning, exercising, and many more - supporting actionable lifestyle insights \cite{vaizman2017context}.




\section{Limitations and future work}

While \modelname{} demonstrates significant advancements in on-device human activity recognition, our approach has several limitations that present opportunities for future research. 


First, watchOS offers limited access to microphone sampling rate adjustments and fails to provide fine-grained battery usage metrics for third-party apps. In practice, this means we cannot precisely adjust the sampling frequency based on energy availability. Addressing these constraints may require kernel-level modifications or leveraging the latest hardware releases, which might include more efficient microphone interfaces or battery optimization APIs.

In addition, although the current \modelname{} model is effective, it may not be fully optimized for the computational constraints of smartwatch hardware. Users with older devices may experience slower inference and increased battery drain due to limited processing power, while newer models often include hardware accelerators -- such as Neural Processing Units (NPUs) in recent Apple devices -- that can significantly improve energy efficiency when properly leveraged. In practical scenarios, energy efficiency could be improved by enabling the model to reduce sampling rates, generate fewer predictions, or skip sensor readings during periods of low activity, such as sitting or resting.

Finally -- and most importantly -- our current evaluation lacks a longitudinal study to assess the model’s performance over extended periods and in diverse real-world situations. Similar to prior works, our accuracies see a severe performance degradation going from semi-controlled datasets to in-the-wild datasets. Since the model can now run on-device, future work should plan passive data collection studies over multiple days—or even weeks or months. A user study focused on long-term use would not only validate the system’s sustained accuracy but also reveal new opportunities for dataset collection, applications, and personalization. 
\section{Safe and Responsible Innovation Statement}

WatchHAR performs all inference on‑device, so raw audio and IMU stay on the watch, eliminating server‑side leakage risks. We train and evaluate solely on previously published, consented datasets (SAMoSA, Semi‑Naturalistic, In‑the‑Wild) and will release code, weights, and evaluation scripts to foster reproducibility. Because dataset demographics are limited, we report per‑dataset results to reveal potential bias and encourage future work on broader populations. Public releases must preserve our privacy‑by‑design constraint: no data storage or export for secondary purposes. We see no foreseeable harms but will address any reports of misuse promptly.

\section{Conclusion}

\modelname{} represents a significant leap forward in on-device human activity recognition on smartwatches for activities of daily living. By successfully implementing a multimodal system that processes both IMU and audio data entirely on-device with a novel end-to-end differentiable preprocessing plus inference pipeline, we have addressed key challenges in privacy, latency, and power efficiency that have long hindered the widespread adoption of continuous activity tracking. Our system's ability to maintain high accuracy across a diverse range of activities while operating in real-time on commodity smartwatch hardware demonstrates the viability of edge-based HAR solutions. Our implementation and demo application are openly available at \url{https://github.com/SPICExLAB/WatchHAR}, paving the way for further research and development in this field, potentially leading to new applications in health monitoring, context-aware computing, and personal analytics.

\begin{acks}
We thank Chenfeng (Jesse) Gao for helping film the video and providing feedback on our figures. Vasco Xu's and Henry Hoffmann's work on this project is supported by NSF (CCF-2119184 CNS-2313190 CCF-1822949 CNS-1956180).
\end{acks}

\bibliographystyle{ACM-Reference-Format}
\bibliography{chi-paper-base}


\end{document}